%% file: acl2018.tex
\documentclass[11pt,a4paper]{article}
\usepackage[hyperref]{acl2018}
\usepackage{times}
\usepackage{latexsym}
\usepackage{ulem}

\usepackage{url}

\aclfinalcopy 

\usepackage{amsmath}
\usepackage{multirow}
\usepackage{graphicx}
\usepackage{url}
\usepackage{subcaption,siunitx,booktabs}

\title{Towards a Better Metric for Evaluating Question Generation Systems}

\author{Preksha Nema$^{\dagger\ddagger}$
\hspace{0.2cm} Mitesh M. Khapra$^{\dagger\ddagger}$\\
  $^\dagger$IIT Madras, India \\
    $^{\ddagger}$ Robert Bosch Center for Data Science and Artificial Intelligence, IIT Madras \\
  {\tt \{preksha,miteshk\}@cse.iitm.ac.in}
}
\date{}

\begin{document}
\maketitle


\input{abstract}
\input{introduction1}
\input{related_work}
\input{current_evaluation}

\section{Human Judgments For Answerability}
\label{human_evaluation}
As mentioned earlier, for AQG, in addition to $n$-gram similarity, we also need to focus on the \textit{answerability} of the generated questions. As illustrated in Section \ref{sec:introduction}, answerability of a question depends on whether it contains all relevant information, such as question type (Wh-types), named entities and content words (often relations). Further, depending on the task (document QA, knowledge-base QA or visual QA) the importance of these words may vary. We perform human evaluations to ascertain the importance of each of these components across different QA tasks. These evaluations allow us to independently analyze the importance of each of these components for the $3$ QA tasks. In the remainder of this section, we describe the (i) process of creating noisy questions (ii) instructions given to the evaluators and the (iii) inferences drawn from human evaluations.

\begin{table*}
    \centering
    \scalebox{0.7}{
    \begin{tabular}{|l|p{10cm}|p{10cm}|}
        \hline
         \textbf{Rating}& \textbf{Description} & \textbf{Examples}  \\\hline
         1 & All important information is missing and it is impossible to answer the question  & ``What is against the \textcolor{blue}{\sout{sign}} ?'', ``Why is using \textcolor{blue}{\sout{O2}}  instead of \textcolor{blue}{\sout{CO2}} less efficient?''\\ \hline
        2 & Most of the important information is missing and I can't infer the answer to the question & ``Which films did \textcolor{blue}{\sout{Lee H. Katzin}} direct ?'', ``Low doses of \textcolor{blue}{\sout{anti-inflammatories}} are sometimes used with what \textcolor{blue}{\sout{classes}} of drugs?''\\ \hline
        3 & Some important information is missing leading to multiple answers & ``What Harvard Alumni \textcolor{blue}{\sout{was the}} Palestine Prime Minister?'', ``What country \textcolor{blue}{\sout{is the}}  teaching subject discussing?'' \\ \hline
        4 & Most of the important information is present and I can infer the answer & ``How \textcolor{blue}{\sout{far}} from the Yard is the Quad located?'',``what \textcolor{blue}{\sout{films}} did Melvin Van Peebles star in?'' \\ \hline
        5 & All important information is present and I can answer the question  & ``What globally popular half marathon began \textcolor{blue}{\sout{in}} 1981?'', ``What kind \textcolor{blue}{\sout{of }} vehicle \textcolor{blue}{\sout{is}} parked \textcolor{blue}{\sout{the}} sidewalk?'' \\
        \hline
    \end{tabular}}
    \caption{Instructions along with the examples. The striked out words were removed as a part of systematic noise from the original question. }
    \label{tab:instructions}
\end{table*}

\subsection{Creating Noisy Questions}
\label{nosiy_dataset}
We took $1000$ questions each from $3$ popular QA datasets, \textit{viz.}, SQuAD, WikiMovies, and VQA. SQuAD \cite{datasquad} is a reading comprehension dataset consisting of around $100$K questions based on passages from around $500$ Wikipedia articles. The WikiMovies dataset contains around $100$K questions which can be answered from a movie knowledge graph containing $43$K entities and $9$ relations \textit{(director, writer, actor, \textit{etc.})}. The VQA dataset is an image QA dataset containing $265,016$ images with around $5.4$ questions on average per image. 

We then created noisy versions of these questions using one of the following four methods:

 \textbf{Dropping function words:} We refer to the list of English function words as defined in NLTK \cite{nltk} and drop all such words from the question. Note that a noisy question with all function words dropped will have a very low BLEU score compared to the original question.

 \textbf{Dropping Named Entities:} In our setup, identifying named entities in questions was easy because the questions were well formed and all named entities were capitalized. Alternately, we could have used the Stanford NER. However, on manual inspection, we found that marking the capitalized words as named entities were sufficient. We randomly dropped at most three named entities per question. This allows us to study how humans rate the output of an AQG system which does not contain the correct named entities.

 \textbf{Dropping Content Words:} Words other than function words and named entities are also crucial for \textit{answerability}. For example, ``Who killed Jane?" and ``Who married Jane?" lead to totally different answers. The word ``killed/married" is very relevant to ascertain the correct answer. These words typically capture the relation between the entities involved (for example, \textit{killed (John, Jane)}). We identify such important (content) words as ones which are neither question types (7-Wh questions) nor named entities nor stop-words. This perturbation allows us to study how humans rate an AQG system which does not produce the correct content (relation) words.

 \textbf{Changing the Question type:} Changing the question type can lead to a different answer altogether or can make the question incoherent. For example the answers to ``Who killed Jane?'' and ``What killed Jane?'' are completely different. We create a noisy question by randomly changing the type of the question (for example. replace ``who'' with ``what''). These question types are well defined ($7$-Wh questions including ``\textit{how}'') and hence it is easy to identify and replace them. This allows us to study the importance of correct question type in the output of an AQG system.

Note that, an alternate way of collecting human judgments would have been to take the output of existing AQG systems and ask humans to assign answerability scores to these questions based on the presence/absence of the above mentioned relevant information. However, when we asked human evaluators to analyze $200$ questions generated by an existing AQG system, they reported that the quality was poor. In particular, after having discussions with annotators, we found that using this output, it would be very difficult to conduct such a systematic study to assess the importance of different words in the question. Hence, we chose to use systematically simulated noisy questions.

\subsection{Instructions}

\begin{table}
    \centering
    \begin{tabular}{|c|c|c|c|}
    \hline
          \textbf{Dataset} & $\kappa$ & Pearson & Spearman \\
          \hline
         SQuAD & 0.63 & 0.823 & 0.795 \\
         \hline
         WikiMovies & 0.81 & 0.934 & 0.927\\
         \hline
          VQA & 0.70 & 0.842 & 0.822\\
         \hline
    \end{tabular}
    \caption{Inter annotator agreement, Pearson and Spearman coefficients between Human Scores.}
    \label{tab:hh_correlation}
\end{table}
We asked the annotators to rate the \textit{answerability} of the above noisy questions on a scale of $1$-$5$. The annotators were clearly told whether the questions belonged to documents or knowledge bases or images. In our initial evaluations, we also tried showing the actual source (image or document) to the annotators. However, we realized that this did not allow us to do an unbiased evaluation of the quality of the questions. The annotators inferred missing information from the document or image and marked the question as answerable (even though the relevant entity \textit{cat} is missing in the question). For example, consider the image of a cat drinking milk and the question ``What is the drinking ?'' If a human is shown the image then she can easily infer that the missing information is ``cat'' and hence mark the question as answerable. This clearly biases the study, and therefore we did not show the source to the evaluators. 

A total of $25$ in-house annotators participated in our study, and we got each question evaluated by $two$ annotators. The annotators were Computer Science graduates competent in English. We did an initial pilot using the instructions mentioned in Table \ref{tab:instructions}, but due to the subjective nature of the task, it was difficult for the annotators to agree on the notion of \textit{important information}. In particular, we found that the annotators disagreed between \textit{most important information} and \textit{all important information} (\textit{i.e.}, they were confused between rating $1$ v/s $2$ and $4$ v/s $5$). We, therefore, did a small pilot with a group of 10 annotators and asked them to evaluate around 30 questions from each dataset and help us refine the guidelines to define the notion of importance clearly.  
Based on group discussions with the annotators we arrived at additional example based guidelines to help them distinguish between cases where \textit{``all the''}, \textit{``most of the''} and \textit{``some of the''} important information is present. 
The original instructions and various examples (some of which are shown in Table \ref{tab:instructions}) were then shared and explained to all the annotators, and they used these to provide their judgments.

\begin{table}
\centering
\resizebox{\columnwidth}{!}{%
\begin{tabular}{|c|rr|rr|rr|}
\hline
\multirow{2}{*}{\textbf{Metric}} & \multicolumn{2}{c|}{\textbf{SQuAD}}                                           & \multicolumn{2}{c|}{\textbf{WikiMovies}}                                      & \multicolumn{2}{c|}{\textbf{VQA}}                                            \\ \cline{2-7} 
                                 & \multicolumn{1}{c}{\textit{Pearson}} & \multicolumn{1}{c|}{\textit{Spearman}} & \multicolumn{1}{c}{\textit{Pearson}} & \multicolumn{1}{c|}{\textit{Spearman}} & \multicolumn{1}{c}{\textit{Pearson}} & \multicolumn{1}{c|}{\textit{Spearman}} \\ \hline
BLEU1                           & 0.167                            & 0.165                                  & 0.179                              & 0.144                               & -0.025*                               & -0.048*                                 \\
BLEU2                           & 0.100*                                & 0.103*                               & 0.072*                            & 0.087*                             & -0.075*                                & -0.091*                                 \\
BLEU3                           & 0.080*                              & 0.086*                                 & 0.036*                               & 0.001*                                & -0.126                                & -0.114                               \\
BLEU4                           & 0.065*                               & 0.067*                                 & -0.020*                             & -0.011*                                 & -0.086*                                & -0.127                            \\
ROUGE-L                          & 0.165                             & 0.158                                   & 0.091*                               & 0.043*                               &-0.009*                               & -0.053*                                \\
METEOR                           & 0.107                             & 0.124                                    & 0.198                             & 0.214                            & -0.035*                               & 0.009*                                 \\
NIST                             & 0.173                                & 0.158                                  & 0.088*                              & -0.033*                                 & 0.158                               & 0.169                                                  \\ \hline
\end{tabular}%
}
\caption{Correlation between existing metrics and human judgments.  Note that the values with $*$ are \textbf{not} statistically significant (p-value $> 0.01$).}
\label{tab:hb}
\end{table}

\begin{figure*}
    \centering
    \resizebox{\textwidth}{!}{%
    \includegraphics[width=0.3\textwidth]{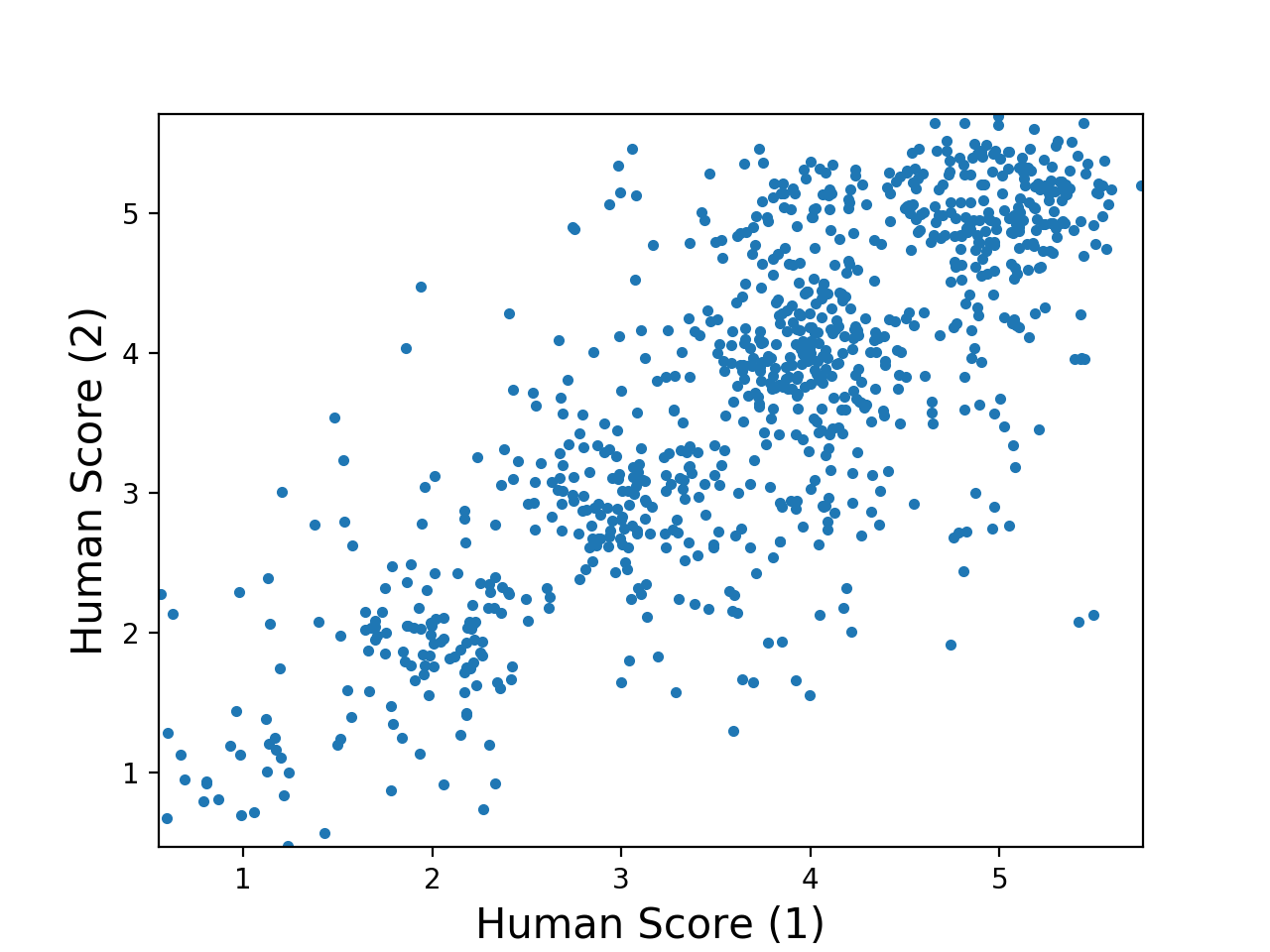}\hfill
    \includegraphics[width=0.3\textwidth]{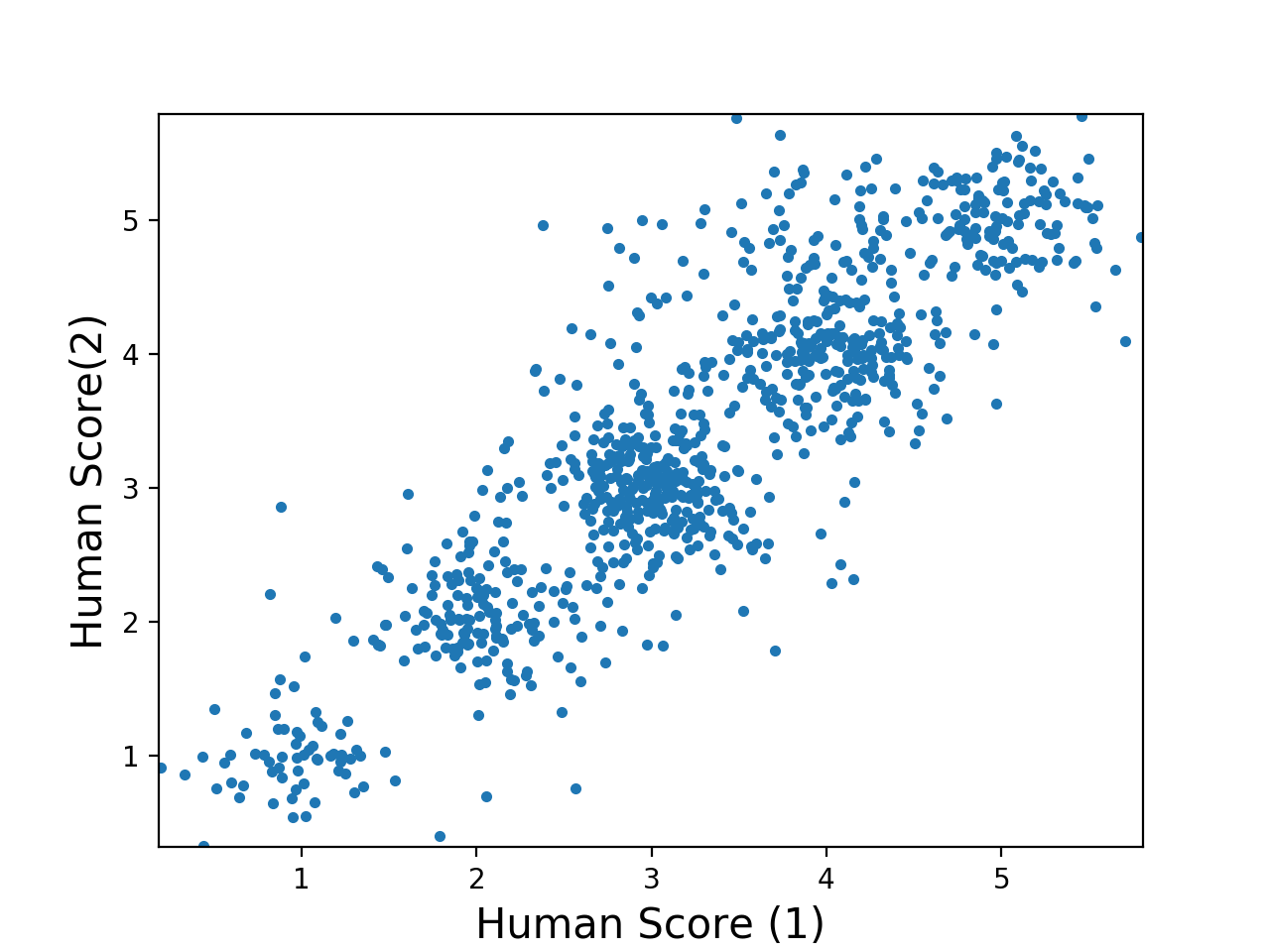}\hfill
    \includegraphics[width=0.3\textwidth]{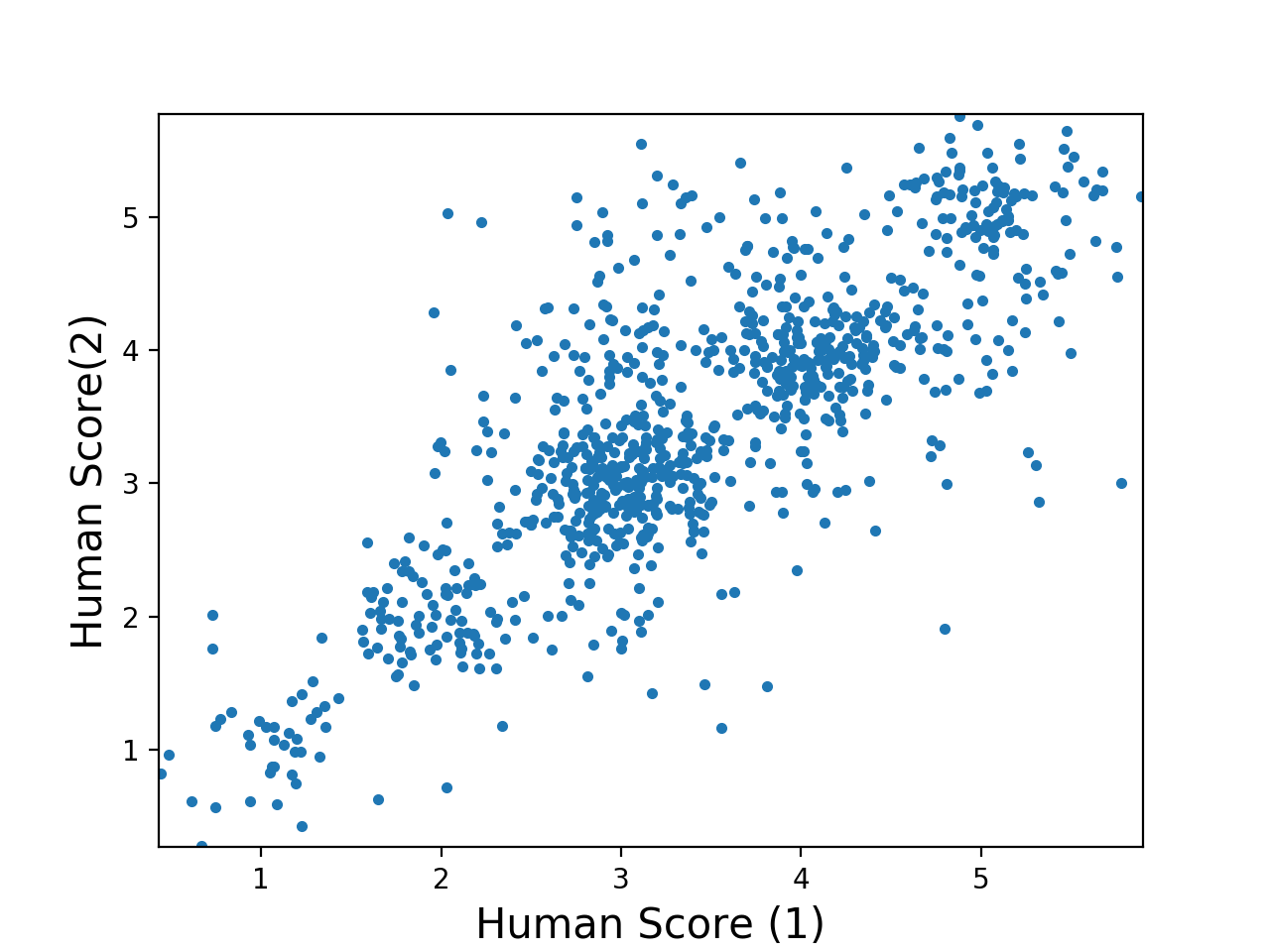}%
    }
\caption{Human-Human Correlation for SQUAD, WikiMovies and VQA respectively.}
\label{fig:hhcor}
\end{figure*}

\subsection{Human-Human Correlation}
In Table \ref{tab:hh_correlation}, we report the average inter-annotator agreement between the ratings using Cohen's kappa ($\kappa$) score \cite{kappa}. Based on guidelines in \cite{interrater} we note that we have a strong inter-annotator agreement for WikiMovies and moderate agreement for SQuAD and VQA. Figure \ref{fig:hhcor} indicates that there is a linear correlation between the two ratings for each question and hence we measured the correlation using Pearson coefficient. For completeness, we also measure the monotonic correlation using Spearman coefficient. The Spearman coefficient is slightly lower than the Pearson coefficient because the inter-annotator agreement is stronger at the tail of the distribution \textit{i.e.}, when the question is either very bad (Rating: $1$) or very good (Rating: $5$).

\subsection{Correlation between human scores and existing evaluation metrics}
We first compute BLEU, METEOR, NIST and ROUGE-L score for each noisy question by comparing it to the original question. We then compute the correlation of each of these scores with annotator ratings. Note that to compute correlation, the annotator ratings are combined to obtain a gold score. The ratings are normalized using the normalization method mentioned in \cite{blatz} and then averaged to obtain the gold score. 
For SQuAD and VQA, we observe that NIST which gives more weightage to informative n-grams correlates better than other metrics. For WikiMovies, METEOR  which even allows non-exact word matches correlates better than other metrics. For SQuAD and WikiMovies, the correlation of human scores with the simple unigram based BLEU1 score is higher than that with other metrics. This is in line with the observation we made earlier that humans can understand and answer questions that are not well-formed, \textit{e.g.}, ``What birth-date Damon?''.

\section{Modifying existing metrics for AQG}
The above study suggests that existing metrics do not correlate well with human judgments about \textit{answerability}. We propose modifications to these metrics so that in addition to $n$-gram similarity they also account for \textit{answerability}. Based on the human evaluations, we found that \textit{answerability} mainly depends on the presence of $4$ types of elements, \textit{viz.}, \textit{relevant content words}, \textit{named entities} and \textit{question types} and \textit{function words}. As outlined in Section \ref{nosiy_dataset} it is easy to identify these elements in the question. Let $c(S_r), c(S_n), c(S_q)$ and $c(S_f)$ be the number of \textbf{r}elevant words, \textbf{n}amed entities, \textbf{q}uestion words and \textbf{f}unction words respectively in the noisy question which have corresponding matching words in the gold standard reference question. We can then compute the weighted average of the precision and recall of each of these elements as

\begin{align*}
    P_{avg} &= \sum_i w_i \frac{c(S_i)}{|l_i|} ~~&~~ R_{avg} &= \sum_i w_i \frac{c(S_i)}{|r_i|}
\end{align*}
where $i\in\{r,n,q,f\}$, $\sum_i w_i = 1$ and $|l_i|$ , $|r_i|$ is the number of the words belonging to $i^{th}$ type of element in the noisy question and reference sentences respectively. Just to be clear $r,n,q,f$ stand for \textit{relevant content words}, \textit{named entities} and \textit{question types} and \textit{function words} respectively. Note that $w_i$'s are tunable weights and in Section \ref{subsec:tune}, we explain how to tune these weights. 

\begin{align*}
            \text{Answerability} &= 2.\frac{P_{avg}R_{avg}}{P_{avg}+R_{avg}}
\end{align*}

We can combine this \textit{answerability} score with any existing metric (say, BLEU4) to derive a modified metric for AQG as shown below:

\begin{equation}
Q\textit{-}\text{BLEU4} = \delta \text{Answerability} + (1-\delta) \text{BLEU4}
\label{eq:qmetric}
\end{equation}

such that $\delta \in \{0,1\}$ to make sure that \textit{Q}-Metric ranges between $0$ to $1$. Similarly, we can derive $Q$-NIST, $Q$-METEOR and so on.


\begin{table}
\centering
\resizebox{\columnwidth}{!}{
\begin{tabular}{|c|l|l|l|l|l|}
\hline
\textbf{Datasets}                & $w_{ner}$                                   & $w_{imp}$                                  & $w_{sw}$                                   & $w_{qt}$   & $\delta$                            \\ \hline

\textbf{SQuAD}                & 0.41                                   & 0.36                                     & 0.03     &  0.20                            & 0.66                               \\ \hline
\textbf{WikiMovies}                     & 0.55                                   & 0.31                                  & 0.02         &0.11                          & 0.83                               \\ \hline
\textbf{VQA}                       & 0.04                                   & 0.59                                  & 0.15             &0.21                      & 0.75                               \\ \hline
\end{tabular}
}
\caption{Coefficients learnt for $Q$-BLEU1 from human judgments across different datasets.}
\label{params_learnt}
\end{table}

\begin{table*}
\centering

\begin{tabular}{|c|rr|rr|rr|}
\hline
\multirow{2}{*}{\textbf{Q-Metric}} & \multicolumn{2}{c|}{\textbf{SQuAD}}                                           & \multicolumn{2}{c|}{\textbf{WikiMovies}}                                      & \multicolumn{2}{c|}{\textbf{VQA}}                                             \\ \cline{2-7} 
                                   & \multicolumn{1}{c}{\textit{Pearson}} & \multicolumn{1}{c|}{\textit{Spearman}} & \multicolumn{1}{c}{\textit{Pearson}} & \multicolumn{1}{c|}{\textit{Spearman}} & \multicolumn{1}{c}{\textit{Pearson}} & \multicolumn{1}{c|}{\textit{Spearman}} \\ \hline
\textit{Q}-BLEU1                           & 0.258                                & 0.255                                  & 0.828                                & 0.841                                  & 0.405                                & 0.384                                  \\
\textit{Q}-BLEU2                           & 0.244                                & 0.243                                  & 0.825                                & 0.835                                  & 0.390                                & 0.360                                  \\
\textit{Q}-BLEU3                           & 0.239                                & 0.240                                  & 0.824                                & 0.837                                  & 0.374                                & 0.331                                  \\
\textit{Q}-BLEU4                           & 0.233                                & 0.232                                  & 0.826                                & 0.837                                  & 0.373                                & 0.311                                  \\
\textit{Q}-ROUGE-L                          & 0.253                                & 0.249                                  & 0.821                                & 0.841                                  & 0.402                                & 0.385                                  \\
\textit{Q}-METEOR                           & 0.158                                & 0.157                                  & 0.821                                & 0.837                                  & 0.402                                & 0.378                                  \\
\textit{Q}-NIST                             & 0.246                                & 0.248                                  & 0.824                                & 0.845                                  & 0.384                                & 0.346                                  \\ \hline
\end{tabular}%

\caption{Correlation between proposed Q-Metric and human judgments. All the correlations have a p-value $< 0.01$ and hence statistically significant.}
\label{tab:hq}
\end{table*}
\subsection{Tuning the weights $w_i$'s and $\delta$}
\label{subsec:tune}
We tuned the weights ($w_i$'s and $\delta$) using the human annotation data. For each source (document, knowledge-base, and images), annotators evaluated $1000$ noisy questions. The annotator scores were first scaled between $0$ to $1$ using the normalization method in \cite{blatz}, and the normalized scores were averaged to obtain the final gold score. For each source, we used $300$ of these annotations and used bagging to find the optimal weights. In particular, we drew $200$ samples randomly from the given set of $300$ samples and did a grid search to find $w_i$'s and $\delta$ such that the $Q$-METRIC computed using Equation \ref{eq:qmetric} had maximum correlation with human scores. We  repeated this process for $k = 20$ times and computed the optimal $w_i$'s and $\delta$ each time. We found that for any given weight ($w_i$) the standard deviation was very low across these $k$ experiments. For each $w_i$ and $\delta$ we obtained the final value by taking an average of the values learned in each of the $k$ experiments. We also observed that the weights did not change much even when we used more data for tuning. Also note that we tuned these weights separately for each metric (\textit{i.e.}, $Q$-BLEU4, $Q$-NIST, $Q$-METEOR and so on). For illustration, we report these weights for $Q$-BLEU1 metric in Table \ref{params_learnt}. As expected, the weights depend on the source from which the question was generated. Note that for WikiMovies, named entities have the highest weight. For VQA content words are most important, as they provide information about the entity being referred to in the question. Note that for SQuAD and VQA, the original base metric also gets weightage comparable to other components, indicating that a fluent question makes it easier to understand thus making it answerable. The overall trend for the values of $w_i$'s was similar for other $Q$-METRICs also (\textit{i.e.}, for $Q$-NIST, $Q$-METEOR and so on).   

\subsection{Correlation between Human scores and different $Q$-METRICs}
Once the weights are tuned, we fix these weights and compute the $Q$-METRIC for the remaining $600$-$700$ examples and report the correlation with human judgments for the same set of examples (see Table \ref{tab:hq}). For a fair comparison, the correlation scores reported in Table \ref{tab:hb} are also on the same $600$-$700$ examples. The correlation scores obtained for different $Q$-METRICs are indeed encouraging.  In particular, we observe that while the correlation of existing metrics with noisy questions generated was very low (Table \ref{tab:hb}),  the correlation of the modified metrics is much higher. This suggests that adding the learnable component for \textit{answerability} and tuning its weights indeed leads to a better-correlated metric. Note that for VQA and SQuAD the correlations are not as high as human-human correlations, but the correlations are still statistically significant. We acknowledge that there is clearly scope for further improvement and the proposed metric is perhaps only a first step towards designing an appropriate metric for AQG. Hopefully, the human evaluation data released as a part of this work will help to design even better metrics for AQG. 



\subsection{Qualitative Analysis}
We have listed some examples in Table \ref{tab:posnegsamples}, which highlight some strengths and weakness of the proposed $Q$-METRIC. We categorize examples as positive/negative depending on the similarity between human scores for answerability and the $Q$-BLEU score. For the examples marked as positive, the $Q$-BLEU score is very close to the \textit{answerability} score given by humans. 

\begin{table*}
\centering
\resizebox{\textwidth}{!}{%
\begin{tabular}{|c|l|l|l|l|l|}
\hline
\multicolumn{2}{|c|}{\textbf{Dataset}}                                                   & \multicolumn{1}{c|}{\textbf{Original  Question}}                                               & \multicolumn{1}{c|}{\textbf{Modified Question}}                                           & \textbf{\begin{tabular}[c]{@{}l@{}}Human\\ Scores\end{tabular}} & \textbf{QBLEU}  \\ \hline
\multicolumn{1}{|l|}{\multirow{4}{*}{\textit{SQuAD}}}                     & \multirow{2}{*}{\textbf{Positive}} & \begin{tabular}[c]{@{}l@{}}What is another type of accountant other than a CPA?\\\end{tabular} & \begin{tabular}[c]{@{}l@{}}What is another type of accountant other than a ?\\\end{tabular} & 0.10                                                          & 0.47         \\
                                                    &                                    & In addition to schools, where else is popularly based authority effective?                                          & In addition schools, where else popularly based authority effective?                                           & 0.85                                                           & 0.83        \\ \cline{2-6} 
                                                    & \multirow{2}{*}{\textbf{Negative}} & When did Tesla begin working for the Continental Edison Company?                                                                  & When did begin working for the Continental Edison Company?                                                           & 0.10                                                           & 0.84            \\
                                                    &                                    &What famous person congratulated him?                                                      & What person congratulated him?                                                                      & 0.85                                                           & 0.17          \\ \hline
\multicolumn{1}{|l|}{\multirow{4}{*}{\textit{VQA}}} & \multirow{2}{*}{\textbf{Positive}}  & What color is the monster truck?                                                                    & What color monster truck?                                                                      & 0.92                                                           & 0.81           \\
\multicolumn{1}{|l|}{}                              &     & What is in the polythene ?                                                                     & What is in the ?                                                                          & 0.10                                                           & 0.14                                           \\ \cline{2-6} 
\multicolumn{1}{|l|}{}                              & \multirow{2}{*}{\textbf{Negative}}                                                                         &   Why there are no leaves on the tree?        &  Why are leaves the tree?                                             & 0.35                                                           & 0.73                \\
\multicolumn{1}{|l|}{}   && How are the carrots prepared in the plate?                                                     & How carrots prepared plate?                                                               & 0.10                                                           & 0.68             \\ \hline

\multicolumn{1}{|l|}{\multirow{4}{*}{\textit{WikiMovies}}} & \multirow{2}{*}{\textbf{Positive}} & what films does Ralf Harolde appear in ?                                                                   & what films Ralf Harolde appear ?                                                                          & 0.97                                                           & 0.91        \\
\multicolumn{1}{|l|}{}                              &                                    & what is a film directed by Eddie Murphy ?                                                                 &Which a film directed by Eddie Murphy ?                                                                 & 0.91                                                           & 0.88           \\ \cline{2-6} 
\multicolumn{1}{|l|}{}                              & \multirow{2}{*}{\textbf{Negative}} & what films does Gerard Butler appear in ?                                              & how does Gerard Butler appear in ?                                                              & 0.15                                                           & 0.89          \\
\multicolumn{1}{|l|}{}                              &                                    & John Conor Brooke appears in which movies ?                                                                   &   appears in which movies ?                                                    & 0.03                                                           & 0.44     \\ \hline
\end{tabular}%
}
\caption{Human (Gold) and $Q$-Metric scores for some of the examples from  the collected human-evaluation data.} 
\label{tab:posnegsamples}
\end{table*}

\begin{table}[t]
\begin{minipage}{0.5\textwidth}
\centering
\small{
\begin{tabular}{|p{2.5cm}|l|l|l|}
\hline
\textbf{Type of Noise} & \textbf{BLEU} & \textbf{QBLEU} & \textbf{Hit 1} \\ \hline
\textbf{None}      & 100           &      100          & 76.5          \\ \hline
\textbf{Stop Words}            & 25.4            &    84.0            & 75.6         \\ \hline
\textbf{Question Type}           & 74.0          &  79.3              & 73.5          \\ \hline
\textbf{Content Words}        & 29.4          &     64.3           & 54.7          \\ \hline
\textbf{Named Entity}                   & 41.9          &     48.5           & 17.97         \\ \hline
\end{tabular}
\caption{Performance obtained by training on different types of noisy questions (WikiMovies).}
\label{machine_wikimovies}}
\end{minipage}
\begin{minipage}{0.5\textwidth}
\sisetup{table-format=-1.2}
\centering
\small{
\begin{tabular}{|p{2.5cm}|c|c|c|}
\hline
\textbf{Noise}                & \multicolumn{1}{l|}{\textbf{BLEU}} & \multicolumn{1}{l|}{\textbf{QBLEU}} & \multicolumn{1}{l|}{\textbf{F1}} \\ \hline
\textbf{None}                     & 100                                & 100                                    & 76.5                                                                \\ \hline
\textbf{Question Type}                          & 80.1                               & 66.1                                    & 69.0                                                        \\ \hline
\textbf{Stop Words}                           & 24.2                               & 61.0                                    & 70.4                                                       \\ \hline
\textbf{Content Words}                      & 60.7                               &   57.1                                  & 64.1                                                         \\ \hline
\textbf{Named Entity}                                  & 77.0                               &  56.0                                   & 73.8                                                            \\ \hline
\end{tabular}}
\caption{Performance obtained by training on different types of noisy questions (SQuAD).}
\label{adv_squad}
\end{minipage}
\begin{minipage}{0.5\textwidth}
\sisetup{table-format=-1.2}
\centering
\small{
\begin{tabular}{|p{2.5cm}|c|c|c|}
\hline
\textbf{Noise} & \multicolumn{1}{l|}{\textbf{BLEU}} & \multicolumn{1}{l|}{\textbf{QBLEU}} &    \multicolumn{1}{l|}{\textbf{Acc(\%)}} \\ \hline
\textbf{None}      & 100                                & 100                                     & 64.4                                 \\ \hline
\textbf{Content Words}           & 49.4                               &  58.2                      & 60.21 \\ \hline

\textbf{Question Type}           & 63.7                               &  50.9                                                                  & 59.81 \\ \hline
\textbf{Stop Words}            & 10.8                               & 37.7                                    & 57.37                                 \\ \hline
\end{tabular}}
\caption{Performance obtained by training on different types of noisy questions (VQA).}
\label{machine_vqa}
\end{minipage}
\end{table}
\section{Extrinsic evaluation}
So far we have shown that existing metrics do not always correlate well with human judgments and it is possible to design metrics which correlate better with human judgments by including a learnable component to focus on \textit{answerability}. We would now like to propose an extrinsic way of evaluating the usefulness of the proposed metric. The motivation for this extrinsic evaluation comes from the fact that one of the intended purposes of the modified metrics is to use them for training QA systems. Suppose we use a particular metric for evaluating the quality of an AQG system and suppose this metric suggests that the questions generated by this system are poor. We would obviously discard this system and not use the questions generated by it to train a QA system. However, if the metric itself is questionable, then it is possible that the questions were good enough, but the metric was not good to evaluate their quality. To study this effect, we create a noisy version of the training data of SQuAD, WikiMovies, and VQA using the same methods outlined in Section \ref{nosiy_dataset}. We then train a state of the art model for each of these tasks on this noisy data and evaluate the trained model on the original test set of each of these datasets. The models that we considered were \cite{rcmodelbidaf} for SQuAD, \cite{datakbwikimovies} for WikiMovies and \cite{mutan} for VQA.

The results of our experiments are summarized in Table \ref{machine_wikimovies} - \ref{machine_vqa}. The first column for each table shows the manner in which the noisy training data was created. The second column shows the BLEU4 score of the noisy questions when compared to the original reference questions (thus it tells us the perceived quality of these questions under the BLEU4 metric). We consider BLEU4 because of all the current metrics used for AQG it is the most popular. Similarly, the third column tells us the perceived quality of these questions under the $Q$-BLEU4 metric. Ideally, we would want that the performance of the model should correlate better with the perceived quality of the training questions as identified by a given metric. We observe that the general trend is better \textit{w.r.t.} the $Q$-BLEU4 metric than the BLEU4 metric (\textit{i.e.}, in general, higher $Q$-BLEU4 indicates better performance and lower $Q$-BLEU4 indicates poor performance). In particular, notice that BLEU4 gives much importance to stop words, but these words hardly have any influence on the final performance. We believe that such an extrinsic evaluation should also be used while designing better metrics and it would help us get better insights.

\section{Conclusion}
The main aim of this work was to objectively examine the utility of existing metrics for AQG. Specifically, we wanted to see if existing metrics account for the \textit{answerability} of the generated questions. To do so, we took noisy generated questions from three different tasks, \textit{viz.}, document QA, knowledge base QA and visual QA, and showed that the \textit{answerability} scores assigned by humans did not correlate well with existing metrics. Based on these studies, we proposed a modification for existing metrics and showed that with the proposed modification these metrics correlate better with human judgments. The proposed modification involves learnable weights which can be tuned (depending on the source) using the human judgments released as a part of this work. Finally, we propose an extrinsic evaluation with the aim of assessing the end utility of these metrics in selecting good AQG systems for creating training data for QA systems. Though the proposed metric correlates better with human judgments,  there is still scope for improvement especially for document QA and visual QA. As future work, we would like to design better metrics for answerability and check if a non-linear combination of different elements in the Q-Metric leads to better correlation with human judgments.

\section{Acknowledgements}
We would like to thank Google for supporting Preksha Nema through their Google India Ph.D. Fellowship Program. We would also like to express our gratitude to the volunteers who participated in human evaluations.

\bibliography{acl2018}
\bibliographystyle{acl_natbib}

\end{document}

%% file: abstract.tex
\begin{abstract}
There has always been criticism for using $n$-gram based similarity metrics, such as BLEU, NIST, \textit{etc}, for evaluating the performance of NLG systems. However, these metrics continue to remain popular and are recently being used for evaluating the performance of systems which automatically generate questions from documents, knowledge graphs, images, \textit{etc}. Given the rising interest in such automatic question generation (AQG) systems, it is important to objectively examine whether these metrics are suitable for this task. In particular, it is important to verify whether such metrics used for evaluating AQG systems focus on \textit{answerability} of the generated question by preferring questions which contain all relevant information such as question type (Wh-types), entities, relations, \textit{etc}. In this work, we show that current automatic evaluation metrics based on $n$-gram similarity do not always correlate well with human judgments about \textit{answerability} of a question. To alleviate this problem and as a first step towards better evaluation metrics for AQG, we introduce a scoring function to capture \textit{answerability} and show that when this scoring function is integrated with existing metrics, they correlate significantly better with human judgments. The scripts and data developed as a part of this work are made publicly available.\footnote{\url{https://github.com/PrekshaNema25/Answerability-Metric}}

\end{abstract}

%% file: introduction1.tex
\section{Introduction} \label{sec:introduction}
The advent of large scale datasets for document Question Answering (QA) \cite{datasquad,datamarco,trivia,duorcsaha} knowledge base driven QA \cite{datasimpleqa, saha2018complex} and Visual QA \cite{VQA,dataclevr} has enabled the development of end-to-end supervised models for QA. 
However, as is always the case, data-hungry neural network based solutions could benefit from even more training data, especially in specific domains which existing datasets do not cater to. Creating newer datasets for specific domains or augmenting existing datasets with more data is a tedious, time-consuming and expensive process. To alleviate this problem and create even more training data, there is growing interest in developing techniques that can automatically generate questions from a given source, say a document \cite{learningtoask, modelwheretofocus}, knowledge base \cite{datakbgenerationeacl, sarathfactoid}, or image \cite{qgvqa} . We refer to this task as Automatic Question Generation (AQG). For example, given the document in Table \ref{tab:docqa}, the task is to automatically generate a question whose answer is also contained in the document.

\begin{table}
    \centering
    \begin{tabular}{p{7cm}}
        \hline
        \textbf{Document:} In \textbf{1648} before the term ``genocide'' had been coined , the Peace of Westphalia was established to protect ethnic, racial and in some instances religious groups. \\
        \textbf{Possible Question:} In which year was the Peace of Westphalia established ?\\
        \hline
    \end{tabular}
    \caption{A sample question generated by a human.}
    \label{tab:docqa}
\end{table}

Given the practical importance of AQG and its potential to influence research in QA, it is not surprising that there has been prolific work in this field in the past one year itself \cite{qgvae,qgvqa,qggoaloriented,learningtoask,modelqgforqa}. Before this field grows further, it is important that the community critically examines the current evaluation metrics being used for this task. In particular, there is a need to closely examine the utility of existing $n$-gram based similarity metrics such as BLEU \cite{bleu}, METEOR \cite{meteor}, NIST \cite{nist}, \textit{etc.} which have been adopted for this task. This work is a first step in that direction where we propose that apart from $n$-gram similarity, any metric for AQG should also take into account the \textit{answerability} of the generated questions. With the help of a few examples below, we illustrate that \textit{answerability} depends on the presence of relevant information such as question type (Wh-types), entities, relations, \textit{etc}, and it is possible that a generated question has a high BLEU score but is still unanswerable and hence not useful.

To begin with, consider the task of answering questions from a Knowledge Base. Let us assume that the intended (gold standard) question is ``Who was the director of Titanic?'' and two different AQG systems generate the following questions ``S1: director of Titanic?'' and ``S2: Who was the director of?''. Any $n$-gram based evaluation metric would obviously assign a higher score to S2 (BLEU3: $81.9$) than S1 (BLEU3: $36.8$). However, as should be obvious S1 contains all the relevant information, and most humans would be easily able to understand and answer this question. A good evaluation metric should capture this notion of \textit{answerability} and give more importance to relevant words in the question which brings us to the question ``Which words are relevant?''

The above example might give the impression that \textit{named entities} are essential but other words are not. However, this is misleading and may not always be the case. For example, consider these questions over an image: ``Are the cats drinking milk?'' v/s ``How many cats are drinking milk?''. These two questions have very different meaning indicating that even words like \textit{are} and \textit{how} are also crucial. Similarly, consider the task of answering questions from a passage titled ``Matt Damon''. In this case, most humans will be able to answer the question ``What is the birth date of'' even though the named entity is missing given that the passage only talks about ``Matt Damon''. Thus, in some cases, depending on the source (document, knowledge base, image) different portions of the question may be important.

To concretize the intuitions developed with the help of the above examples, we first collect human judgments. Specifically, we take questions from existing datasets for document QA, knowledge base QA and visual QA and add systematic noise to these questions. We show these questions to humans and ask them to assign scores to these questions based on the \textit{answerability} and hence the usefulness of these questions (\textit{i.e.}, whether the question contains enough information for them to be able to answer it correctly). We also compute various $n$-gram similarity metrics (BLEU, METEOR, NIST) comparing the noisy questions to the original questions and show that these metrics do not correlate well with human judgments. Similar studies \cite{bleucritcs, evalhownottoeval} have already shown that these metrics do not correlate well with \textit{fluency, adequacy, coherence} but in this work, we focus on \textit{answerability}. 

Based on the human evaluations, we propose to modify existing metrics to focus on \textit{answerability} in addition to $n$-gram similarity. The idea is to make these metrics flexible such that, if needed, the weight assigned to \textit{answerability} and $n$-gram similarity can be adjusted depending on the task  (document QA, Knowledge-Base QA, Visual QA). 
 Further, for capturing \textit{answerability} we propose additional weights for different components of the question (question type, content words, function words, and named entities) These weights can be learned from a small amount of human annotated data and may differ from task to task. 


%% file: related_work.tex
\section{ Related Work} \label{sec:rw}
We have organized our literature survey into $2$ parts: (i) question generation systems (ii) studies which analyze evaluation metrics used for NLG.

\textbf{Question Generation:} Early work on question generation used rule-based approaches to generate questions from declarative sentences \cite{good:question, DBLP:conf/aied/MostowC09, DBLP:conf/enlg/LindbergPNW13, DBLP:conf/acl/LabutovBV15}. More recent works use attention based neural models 
for question generation \cite{modelwheretofocus,learningtoask}. Some models \cite{modeltexttotext} feed the generated questions to a QA system and use the performance of the QA system as an indicator of the quality of the questions. A few models \cite{modeljointmodelqaqg, modeldualtask} treat question answering (QA) and question generation (QG) as complementary tasks and focus on jointly training for these two tasks. Other models focus only on the performance of the QA task \cite{modelsemisupervised,modelqgforqa} and not explicitly on the quality of the generated questions. Apart from generating questions from text there is also research on generating questions from images \cite{qgvae, qgvqa, qggoaloriented} and knowledge base \cite{sarathfactoid, datakbgenerationeacl}.

 \textbf{Evaluation metrics for NLG:} Current popular metrics for NLG such as BLEU \cite{bleu}, METEOR\cite{meteor}, ROUGE \cite{rouge} and NIST \cite{nist} essentially compute the $n$-gram similarity between the reference sentence and the generated sentence. Though these metrics are very popular and are used for a wide range of NLG tasks including AQG, there has always been criticism for using these metrics (for example, see \cite{bleucritcs,bluesfobleu, bleucritcs2}). More recently, there has been criticism \cite{evalhownottoeval} for using such metrics for evaluating dialog systems eventually resulting in a new metric \cite{evalnewmetricdiaglog}. This new metric while very important, came a bit late in the day and much after several dialog systems were proposed, evaluated and compared using the above $n$-gram based metrics. It is very important to prevent a similar situation in question generation where many systems get proposed without evaluating them using the right metric. Our work is a first step in this direction, and we hope it will lead to more research in designing the right metrics for AQG.

%% file: current_evaluation.tex
\section{Current Evaluation Metrics} \label{current_eval}
We give a quick overview of the metrics which are currently used for evaluating AQG systems. 

\textbf{BLEU:} BLEU is a precision-based evaluation metric which considers exact $n$-gram matches. 
For a given value of $n$, the precision is computed as the fraction of $n$-grams in the generated hypothesis which match some $n$-gram in the reference hypothesis. The final BLEU score is computed as the geometric mean of the $n$-gram precisions obtained by varying $n$ from $1$ to $N$ where $N$ is typically $3$ or $4$. It also contains a brevity penalty to penalize hypothesis that are too short.  

\textbf{METEOR:} As opposed to BLEU, METEOR uses both precision and recall, \textit{i.e.}, it computes the fraction of the hypothesis which matches the reference (precision) as well as the fraction of the reference which is contained in the hypothesis (recall). Further, unlike BLEU which only considers exact matches, METEOR also considers matches with stemmed words, synonyms, and paraphrases. It also gives different weightage to matches corresponding to function words and matches corresponding to content words. The final score is the harmonic mean of the precision and recall calculated based on these four matches. Additionally, it also includes a fragmentation penalty to account for gaps and differences in word order. In effect, METEOR is a parametric metric where the different parameters, \textit{viz.}, (i) fragmentation penalty, (ii) weights of different matchers (exact, stemmed, synonyms, paraphrases) and (iii) weights of function and content words, are tuned to maximize correlation with human judgments.

\textbf{NIST:} NIST is a variant of the standard BLEU metric that takes into account the relative importance of each $n$-grams in the sentence. In particular, the metric gives a high weightage to $n$-grams which have a lower frequency in the corpus and hence are more informative as compared to very frequent $n$-grams which are less informative. Further, unlike BLEU which takes the geometric mean of $n$-gram precisions, NIST takes the arithmetic mean of these precisions. Additionally, they make a small change to the brevity penalty to minimize the impact of minor variations in the length of the hypothesis.

\textbf{ROUGE:} ROUGE is a set of evaluation metrics which were proposed in the context of automatic summarization. Typically, most studies use ROUGE-L, which is F-measure based on the Longest Common Subsequence (LCS) between a candidate and target sentence. Given two sequences, a common subsequence is the set of words which appear in both the sequences in the same order but unlike $n$-grams the common subsequence does not need to be contiguous. LCS is the longest of such common subsequences. For example, given the sentences candidate:``the boy went home'' and reference:``the boy will go home'', ``the boy home'' is the longest common subsequence even though it is not contiguous.


